\documentclass[10pt,twocolumn,letterpaper]{article}
\usepackage[pagenumbers]{cvpr}

\usepackage{graphicx}
\usepackage{amsmath}
\usepackage{amssymb}
\usepackage{booktabs}
\usepackage{multirow}
\usepackage{float}
\usepackage{dsfont}
\usepackage[pagebackref,breaklinks,colorlinks]{hyperref}
\usepackage[accsupp]{axessibility}

\usepackage[capitalize]{cleveref}
\crefname{section}{Sec.}{Secs.}
\Crefname{section}{Section}{Sections}
\Crefname{table}{Table}{Tables}
\crefname{table}{Tab.}{Tabs.}

\begin{document}

\title{Text Spotting Transformers}

\author{Xiang Zhang$^1$ \quad Yongwen Su$^2$ \quad Subarna Tripathi$^3$ \quad Zhuowen Tu$^1$\\
$^1$UC San Diego \quad $^2$Shanghai Jiao Tong University \quad $^3$Intel Labs\\
{\tt\small \{xiz102, ztu\}@ucsd.edu, heyue2001@gmail.com, subarna.tripathi@intel.com}}
\maketitle

\begin{abstract}

In this paper, we present TExt Spotting TRansformers (TESTR), a generic end-to-end text spotting framework using Transformers for text detection and recognition in the wild. TESTR builds upon a single encoder and dual decoders for the joint text-box control point regression and character recognition. Other than most existing literature, our method is free from Region-of-Interest operations and heuristics-driven post-processing procedures; TESTR is particularly effective when dealing with curved text-boxes where special cares are needed for the adaptation of the traditional bounding-box representations.  We show our canonical representation of control points suitable for text instances in both Bezier curve and polygon annotations. In addition, we design a bounding-box guided polygon detection (box-to-polygon) process. Experiments on curved and arbitrarily shaped datasets demonstrate state-of-the-art performances of the proposed TESTR algorithm.
\let\thefootnote\relax\footnotetext{Code available at \url{https://github.com/mlpc-ucsd/TESTR}.}
\end{abstract}

\section{Introduction}
\label{sec:intro}
Text detection and recognition in natural scenes, called text spotting, is an active area of research in computer vision \cite{shi2015endtoend,Li2017TETS,He2018AnET,liu2018fots,lyu2018mask,liu2020abcnet,qiao2021mango}. Text spotting is of great importance in real-world applications such as mapping, autonomous driving, and image retrieval. The text spotting problem typically consists of two sub-tasks: 1) text detection that localizes text boxes in a natural image, and 2) text recognition that reads the characters from the detected text.
Despite its practical significance and a steady progress observed recently, text spotting remains a challenging problem that requires further improvement. The main difficulty in text spotting is contributed by multiple factors including large variations in font, size, style, color, shape, occlusion, distortion, and layout for natural scene images.

Classical text spotting methods \cite{shi2015endtoend,liu2020abcnet} often perform text detection and recognition in two separate steps. In the detection module, the regions of interest are proposed for text instance detection. After alignment, the features are then used in the text recognition module. In natural scenes, text-boxes often appear in arbitrary orientations \cite{yao2012detecting} and are non-rectangular \cite{liu2020abcnet}. This poses further challenges for the algorithm development that typically requires a number of heuristics designs with intermediate and post-processing steps \cite{Li2017TETS,He2018AnET,lyu2018mask,qiao2021text}.

\begin{figure}
    \centering
    \includegraphics[width=\linewidth]{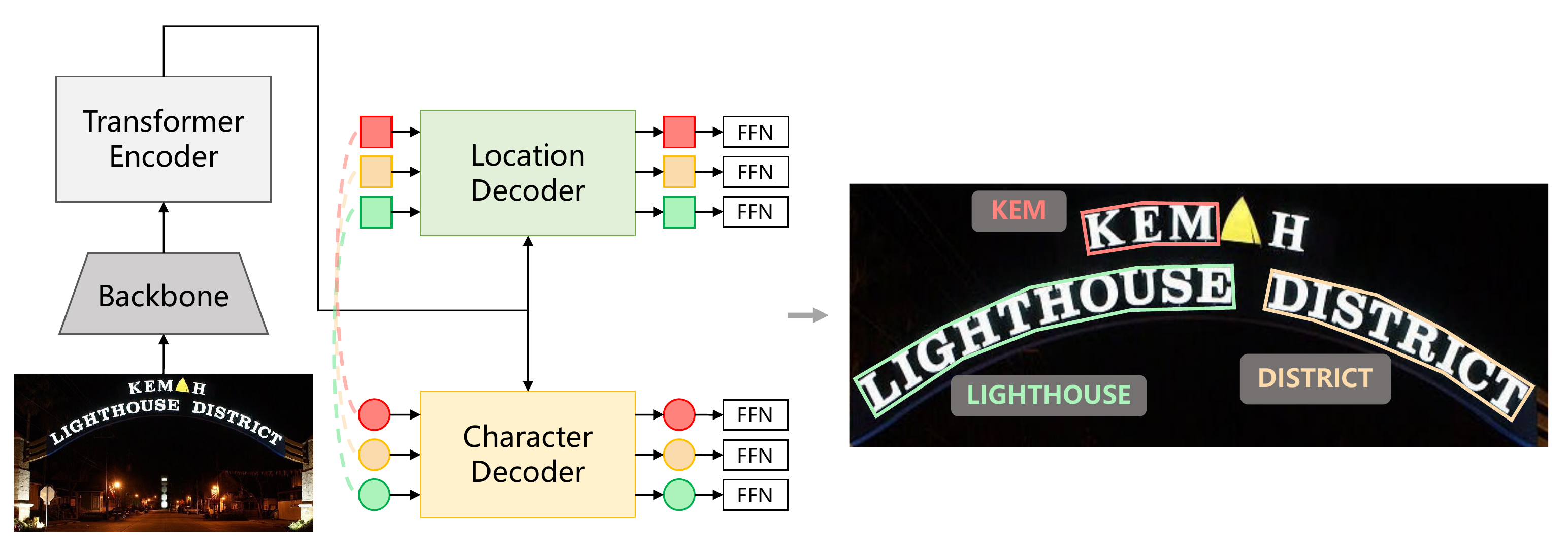}
    \vspace{-1em}
    \caption{\small Illustration of the overall TESTR pipeline. The input image is passed through a feature backbone and Transformer encoder, and the multi-scale feature is shared across the location and character decoder, which predict the coordinates of control points and characters of the text instance respectively. The canonical representation of control points serves both polygon vertices and Bezier curve control points.}
    \label{fig:teaser}
    \vspace{-2mm}
\end{figure}

Transformers \cite{vaswani2017attention} have achieved a remarkable success in natural language processing \cite{devlin2019bert} and computer vision \cite{dosovitskiy2021image}. DEtection TRansformers (DETR) \cite{carion2020end} have also made a profound impact to object detection by removing the proposal anchors and the non-maximum suppression processes needed in the sliding window based approaches \cite{ren2015faster}. LETR \cite{xu2021line} extends DETR by adopting Transformers to directly detect geometric structures such as line segments beyond the bounding box representation.

Inspired by the DETR family models \cite{carion2020end,zhu2021deformable,xu2021line,li2021pose}, we propose TExt Spotting TRansformers (TESTR), a Transformer-based text spotting method that performs text detection and recognition in a unified framework. TESTR avoids the heuristics design and the intermediate stages required in many of the existing text spotting approaches.

\begin{figure*}[!ht]
    \centering
    \includegraphics[width=\linewidth]{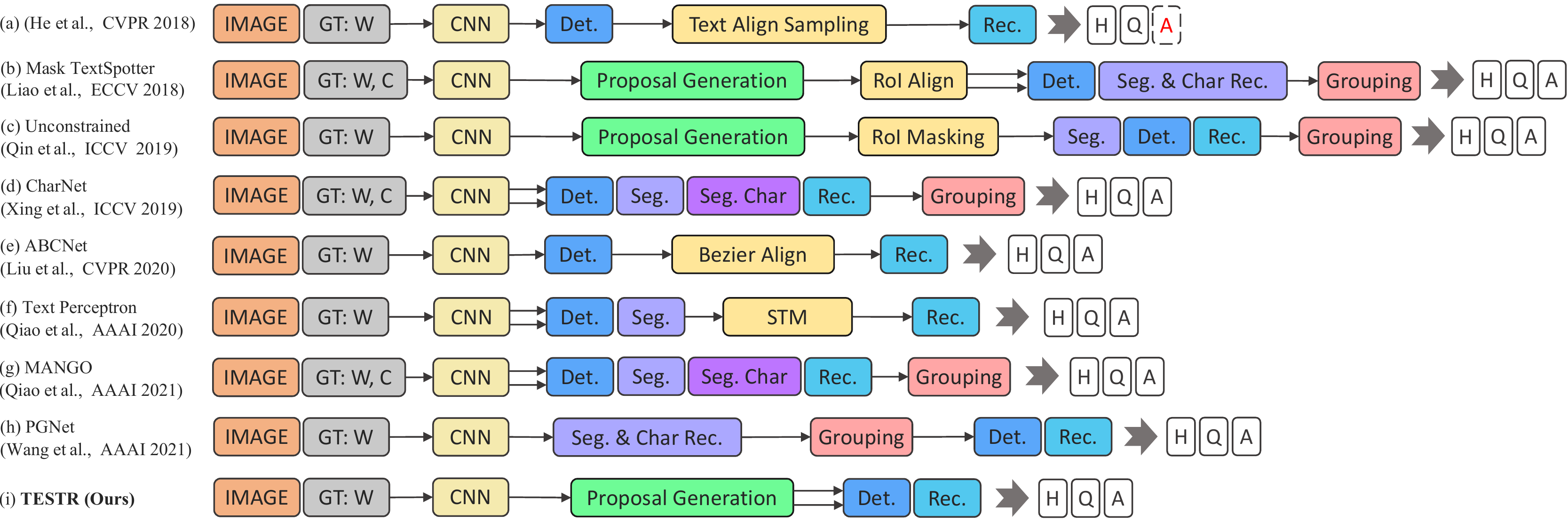}
    \caption{\small Overview of some end-to-end scene text spotting methods that are most relevant to ours. Inside the GT (ground-truth) box, `W' and `C' represent word-level annotation and character-level annotation. The `H', `Q', and `A' represent that the method is able to detect horizontal, quadrilateral, and arbitrarily-shaped text, respectively. The dashed box represents the shape of the text which the method is unable to detect. Figure style from \cite{liu2020abcnet, Wang_2021_pgnet}.}
    \label{fig:related_works}
    \vspace{-1mm}
\end{figure*}

The contribution of TESTR is listed as follows.
\begin{itemize}
    \setlength\itemsep{0.1em}
    \item We propose a single-encoder dual-decoder framework that jointly performs curved text instance detection and recognition using {\bf Transformers beyond the standard bounding box representation}. Our method, thanks to direct regression of the control points coordinates, is a holistic approach that requires neither heuristics-driven post-processing procedures, nor Region-of-Interest operations. 
    \item We introduce a {\bf box-to-polygon process} that achieves bounding-box guided polygon detection in the detection Transformers. Experimental results show an apparent performance boost.
    \item The canonical representation of control points makes our method appropriate for {\bf both the polygonal and Bezier curve} annotations. TESTR achieves state-of-the-art performances on challenging datasets, \ie Total-Text and CTW1500.
\end{itemize}

\begin{figure*}[!ht]
    \centering
    \includegraphics[width=\linewidth]{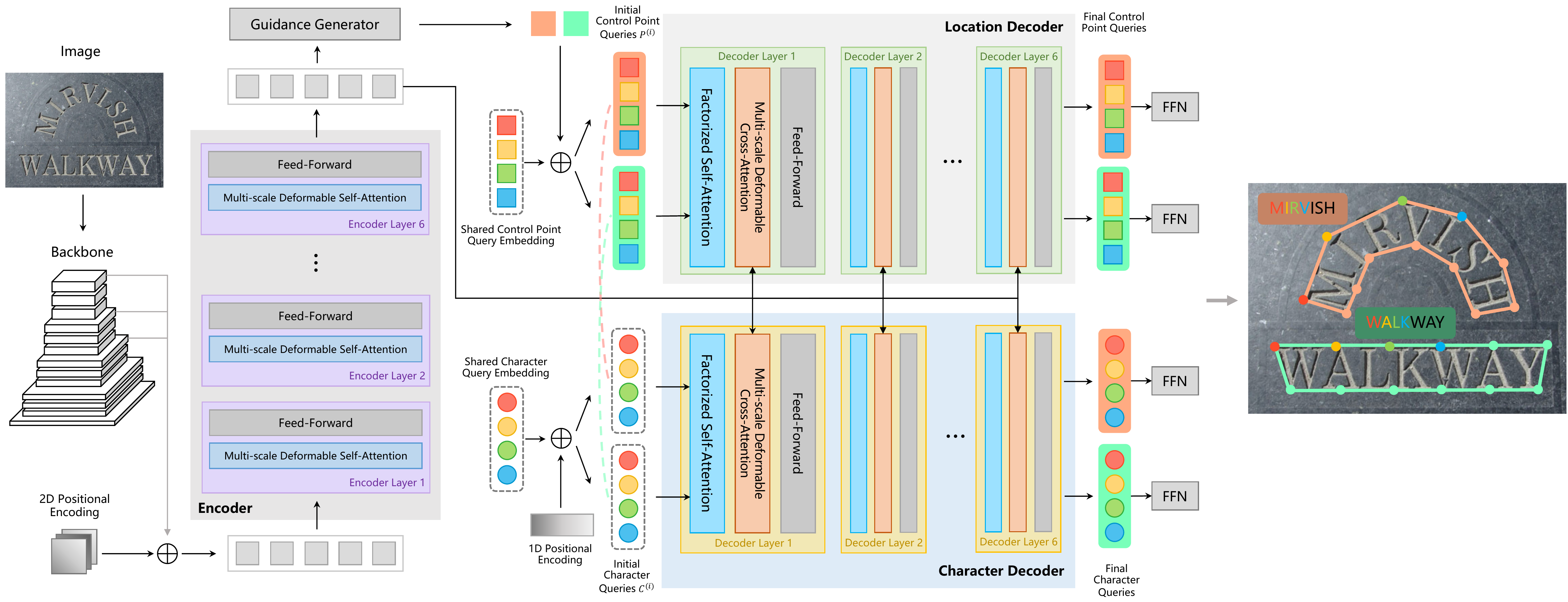}
    \caption{\small Overall architecture of Text Spotting Transformers (TESTR). First, the encoder performs multi-scale deformable self-attention across feature maps, and a guidance generator produces coarse bounding boxes from the features. These boxes are encoded and added on top of the learnable control point query embeddings to guide the learning of control points. Control point queries are fed through the location decoder and feed-forward networks (FFNs) to predict their coordinates. The character decoder, with shared reference points as the location decoder for the multi-scale cross-attention, predicts characters for the corresponding text instance. The framework is end-to-end trainable and performs detection and recognition in a unified way. Note that the control point and character queries with identical background color belong to the same text instance in the output image.}
    \label{fig:main_arch}
    \vspace{-3mm}
\end{figure*}

\subsection{Related Works}

Scene text spotting consists of text detection and recognition. 
Two-stage approaches are first developed to address the task, which train detection and recognition modules separately and simply join them during inference. Recent literature focuses on end-to-end methods, which tackles detection and recognition simultaneously through RoI operations during training. While these methods demonstrate satisfying performance, the text spotting task still remains a challenge due to the prevalence of arbitrarily-shaped texts. We will discuss related works from the perspectives of text detection, text recognition, regular text spotting, and arbitrarily-shaped text spotting. Figure \ref{fig:related_works} is an overview of exemplary works.

\vspace{.5em}

 \textbf{Text Detection.} Early works\cite{epshtein2010detecting,liao2016textboxes,tian2016detecting} focus on horizontal text detection, which predict rectangular bounding boxes for the text instances. They pose apparent limitations as texts in the wild are mostly multi-oriented quadrilateral, curved, or even arbitrarily shaped. Efforts have been made to address these challenging cases. \cite{zhou2017east,liao2018textboxes++} use both rotated boxes and quadrangles to achieve multi-oriented quadrilateral text detection. \cite{baek2019character} enables the detection of arbitrarily-shaped texts through the prediction of character boxes. While achieving a dramatic performance boost, it requires expensive character-level annotations and post-processing to group the detected characters back to texts. \cite{wang2019arbitrary} uses pairwise point representation for text regions, yet it is restricted to the sequential decoding of RNNs. \cite{liu2020abcnet,liu2021abcnet} introduce a novel Bezier curve representation of the curved texts, and significantly improve the detection performance on them.

\vspace{.5em}

\textbf{Text Recognition.} Classical works \cite{wang2011end,mishra2012top,neumann2012real} adopt statistical approaches to classify characters and group them into words. Deep-learning based methods\cite{Su2014TextRNN,lee2016recursive} have ushered a new era for text recognition. CRNN\cite{shi2015endtoend} integrates CNN and RNN to perform text recognition. However, it is mainly applicable to regular texts and limited as to arbitrarily-shaped texts. \cite{shi2016robust,liu2016starnet} use a spatial transformer to convert irregular texts into rectangular shapes, and then feed them into the feature extractor and sequence decoder for recognition.

\vspace{.5em}

\textbf{Regular End-to-end Scene Text Spotting.} To further enhance the performance of text spotting, \cite{Li2017TETS} proposes an end-to-end trainable text spotting framework. RoI Pooling is introduced to bridge the gap between text detection and recognition. However, this method is limited to horizontal texts. Other literature conducts quadrilateral text spotting based on other specially crafted RoI operations, such as Text-Align\cite{He2018AnET} and RoI-Rotate\cite{liu2018fots}, while remaining incapable of spotting arbitrarily-shaped texts.

\textbf{Arbitrarily-shaped End-to-end Scene Text Spotting.}
In \cite{sun2018textnet}, quadrangle text region proposals are generated, followed by an RoI transform. While this method can recognize irregular texts, its quadrilateral representation is not optimal for arbitrarily-shaped text regions. CharNet~\cite{xing2019convolutional} performs character and text detection in a single pass, requiring character-level annotations. TextDragon~\cite{feng2019textdragon} generates multiple local quadrangles around the text centerline, with RoISlide operation for feature warping and aggregation within the text instance. Though not requiring character-level supervision, it still needs to perform centerline detection, grouping, and sorting to convert local quadrangles to text boundaries.

Other literature focuses on segmentation-based methods for arbitrarily-shaped text spotting. Mask TextSpotter\cite{lyu2018mask}, built on Mask R-CNN\cite{he2017mask}, performs text- and character-level segmentation, requiring further grouping before getting final results. \cite{qin2019unconstrained} proposes RoI masking that multiplies segmentation probability maps with features to suppress the background, whereas \cite{liao2020mask} uses binary maps to mitigate the inaccuracies in segmentation. While these approaches achieve fair performance, the mask representation is subject to post-processing such as polygon fitting and smoothing to obtain desirable boundaries. MANGO~\cite{qiao2021mango} develops Mask Attention module to retain global features for multiple instances, yet it still requires centerline segmentation to guide the grouping of the predictions. 

Recent works try to develop appropriate representations that directly capture the text boundaries. ABCNet\cite{liu2020abcnet} and ABCNet v2\cite{liu2021abcnet} introduce parametric Bezier curve representations for curved texts, and develop Bezier-Align for feature extracting. However, low-order Bezier curves exhibit limitations when representing heavily curved or wavy text shapes. \cite{qiao2021text} uses Shape Transform Module to generate fiducial points around the text boundaries and rectify irregular texts. PGNet~\cite{Wang_2021_pgnet} transforms the polygonal text boundaries to the centerline, border offset, and direction offset and perform multi-task learning for these objectives. While eliminating RoI operations, it still uses a specially designed polygon restoration process. 

In contrast, our approach relies solely on Transformers, which is entirely free from RoI operations. With the outputs being coordinates of polygon vertices or Bezier control points for the text instance, along with the corresponding character sequence, no special post-processing is needed. 

\section{Method}

TExt Spotting TRansformers (TESTR) is an end-to-end trainable framework that handles text detection and recognition in a unified manner. The overall architecture is shown in Figure \ref{fig:main_arch}. We first introduce the multi-scale deformable attention as in Deformable DETR~\cite{zhu2021deformable}, and elaborate on the key components of our model -- dual decoders for detection and recognition, and box-to-polygon detection procedure.

\subsection{Multi-Scale Deformable Attention}

One obstacle for the text spotting task is the prevalence of small text instances in the images.
Current literature tries to overcome this limitation by leveraging multi-scale feature maps, such as Feature Pyramid Network (FPN) \cite{Lin_2017_CVPR_fpn}. To utilize such feature maps, we take the multi-scale deformable attention module in \cite{zhu2021deformable}. Given a set of $L$ level multi-scale feature maps $\{U_l\}_{l=1}^L$, with each level as $U_l \in \mathbb{R}^{C\times H_l \times W_l}$, and $\mathbf{p}(q)$ as the normalized coordinates of the reference point for the query $q$, the multi-scale deformable attention can be expressed as 
\begin{equation}
\resizebox{.9\linewidth}{!}{%
$\begin{aligned}
   & \text{MSDeformAttn}(q, \; \mathbf{p}(q), \; \{U_l\}_{l=1}^L) = \\ 
   & \sum_{h=1}^H \mathbf{W}_h \left\{ \sum_{l=1}^L \sum_{k=1}^K \mathbf{A}_{hlk}(q) \cdot \mathbf{W}'_h U_l \left[ \phi_l \left(\mathbf{p}(q) \right)  + \Delta \mathbf{p}_{hlk} (q) \right] \right\}
\end{aligned}$
}
\end{equation}

where $h, \;l, \;k$ are indices for the attention head, input feature level, and sampling point respectively. $\mathbf{A}_{hlk}$ denotes the attention weight for query $q$, normalized with respect to $K$ sampling points. $\phi_l$ maps the normalized coordinates to the scale of $l$-th level feature map, and $\Delta \mathbf{p}$ generates an appropriate sampling offset for the query. Both of them are added to form the sampling location for the feature map $U_l$. $\mathbf{W}_h'$ and $\mathbf{W}_h$ are trainable weight matrices that are similar to those present in the original multi-head attention. 

Instead of relying on the original attention, which requires sampling of $H \times W$ points in the feature map, multi-scale deformable attention samples $L K$ points, largely reducing computational overheads and enabling the capability to use multi-scale feature maps. We will illustrate its efficiency in the experiment section.

\subsection{Dual Decoders}\label{sec:dual_decoders}

We formulate the holistic text spotting task as a set prediction problem. Given an image $I$, we need to output a set of point-character tuples, defined as $Y=\left\{\left( \mathbf{P}^{(i)}, \; \mathbf{C}^{(i)}\right)\right\}_{i=1}^K$, where $i$ is the index for each text instance, $\mathbf{P}^{(i)} = (p_1^{(i)},\;\cdots,\;p_N^{(i)})$ is the coordinates of $N$ control points, and $\mathbf{C}^{(i)} = (c_1^{(i)},\;\cdots,\;c_M^{(i)})$ is the $M$ characters of the text. 

To tackle this problem, we propose a dual-decoder paradigm for predictions of different modalities, location decoder for detection (to predict $\mathbf{P}^{(i)}$) and character decoder for recognition (to predict $\mathbf{C}^{(i)}$).

\vspace{.5em}

\textbf{Location decoder.} We extend the queries in original DETR\cite{carion2020end} to composite queries for predicting multiple control points for each instance. We have $Q$ such queries, each corresponding to a text instance, as $\mathbf{P}^{(i)}$. Each query element is composed of subqueries $p_j$, where $\mathbf{P}^{(i)} = (p_1^{(i)},\;\cdots,\;p_N^{(i)})$. To capture the relationship between different text instances and between different subqueries within a single text instance in a structural way, we utilize factorized self-attention, inspired by \cite{Dong_2021_ICCV}. The factorized self-attention is composed of an intra-group attention, which is a self-attention within subqueries belonging to each of the $\mathbf{P}^{(i)}$, and an inter-group attention, which is a self-attention across $p_j$ of different queries. 

The initial control point queries are fed into the location decoder. After the process of multi-layer decoding, the final control point queries are taken by a classification head predicting the confidence, and a 2-channel regression head outputting the normalized coordinates for each control point. 

The control points predicted here can either be $N$ polygon vertices, or control points for Bezier curves, as in \cite{liu2020abcnet}. For the polygon points, we use the sequence that starts with the top left corner and moves in the clockwise order.

For the Bezier control points, Bernstein Polynomials\cite{lorentz2013bernstein} can be used to construct the parametric curve 
\begin{equation}
    c(t) = \sum_{j=1}^{N} p_j B_{(j-1), (N-1)}(t),\;\; t \in [0,\;1]
\end{equation}

where Bernstein basis polynomials are defined as 
\begin{equation}
    B_{i,n}(t) = \binom{n}{i} t^i(1-t)^{n-i},\;\; i=0,\;1,\;\cdots,\;n 
\end{equation}

Following \cite{liu2020abcnet}, we use two cubic Bezier curves for a single text instance, corresponding to the two possibly curved sides of the text. One can sample across $t$ to convert Bezier curves back to polygons. 

\vspace{.5em}

\textbf{Character decoder.} The character decoder follows most of the location decoder, with control point queries replaced by character queries $\mathbf{C}^{(i)}$. The initial character queries comprise a learnable query embedding and 1D sine positional encoding, and are shared across different text instances. The character query $\mathbf{C}^{(i)}$ and control point query $\mathbf{P}^{(i)}$ with the same index belong to the same text instance, and therefore the reference points of the multi-scale deformable cross-attention are shared to ensure they get the identical contexts from the image feature. A classification head takes the final character queries to predict among multiple character classes. 

\begin{figure}
    \centering
    \includegraphics[width=\linewidth]{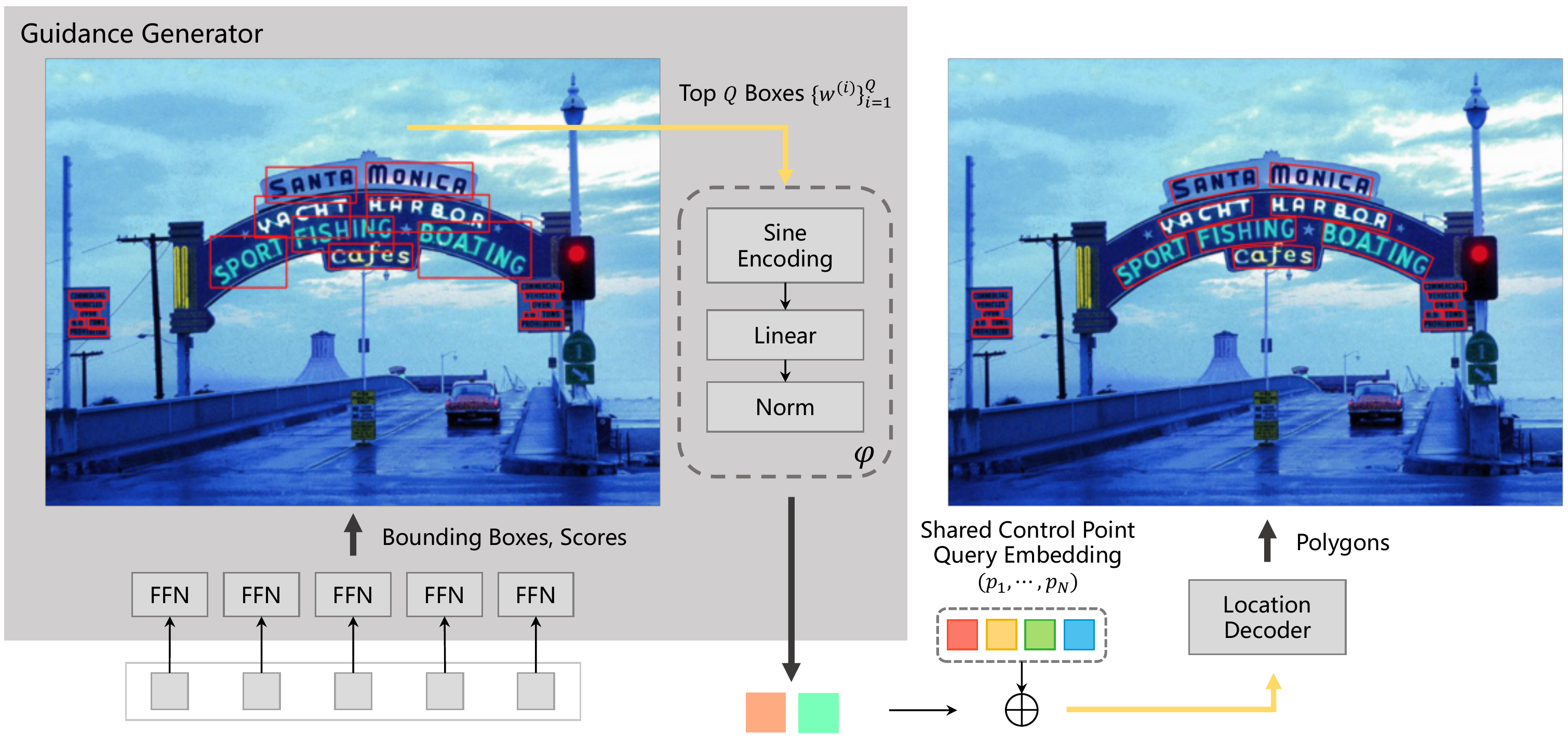}
    \vspace{-1em}
    \caption{\small Illustration of the box-to-polygon detection process. The guidance generator predicts coarse bounding boxes and scores as shown in the left image. The coordinates of top $Q$ boxes are fed to the differentiable encoding module $\varphi$ and the encoded results are added to the shared control point query embeddings, which are taken by the location decoder for the final polygonal predictions.}
    \label{fig:box_to_poly}
    \vspace{-1mm}
\end{figure}

\subsection{Box-to-Polygon Detection Process} \label{sec:box_to_poly}
The decoder models the Bayesian inference process $P(Y | I) \propto P(I | Y) P(Y)$ for our set prediction problem, where $P(I | Y)$ captures the relationship between hypotheses (queries) and input $I$ through cross-attention, while $P(Y)$ models the prior on configuration of $Y$ through self-attention. We argue that when $Y$ is complex, in our case of composite queries, $P(Y)$ is hard to learn. Hence, we propose a box-to-polygon detection approach, which takes the bounding boxes of text instances and uses them to guide the polygon detection. This process, employing information related to concrete image $I$ to form input-specific priors, facilitates the training of polygon control point regression.

The framework begins with the guidance generator in Figure \ref{fig:main_arch}, which is a proposal generator outputting coarse bounding box coordinates and probabilities. Boxes with top-$Q$ probabilities are selected and their coordinates are denoted as $\{ w^{(i)} \}_{i=1}^Q$. The initial control point queries described in \ref{sec:dual_decoders} are formed by:
\begin{equation}
    \mathbf{P}^{(i)} = \varphi (w^{(i)}) + (p_1,\;\cdots,\;p_N) \label{eqn:box_guidance}
\end{equation}
where $(p_1,\;\cdots,\;p_N)$ is the control point query embedding, shared across $Q$ queries, modeling the general relation between control points that is irrelevant to the specific bounding box location. $\varphi$ is the sine positional encoding function followed by a linear and normalization layer, and therefore is fully differentiable. $\varphi(w^{(i)})$, as the encoded bounding box information, is shared across $N$ subqueries within a single instance, modeling the overall location and scale of the text instance. $w^{(i)}$ is also used as the initial reference point for the multi-scale deformable cross-attention.

An illustration of this process is provided in Figure \ref{fig:box_to_poly} with details of the guidance generator. Ablation studies in Section \ref{sec:ablations} demonstrate the significant improvement in the recognition accuracy brought by this process.

\subsection{Training Losses} 

\vspace{.5em}

\textbf{Bipartite matching.} Since TESTR outputs a fixed number of predictions unlike the actual number ($G$) of ground truth instances, we need to find an optimal matching between them to calculate the loss. Specifically, we need to find an injective function $\sigma: [G] \mapsto [Q]$ that minimizes the following matching cost $\mathcal{C}$:
\begin{equation}
    \underset{\sigma}{\arg \min} \sum_{i=1}^{G} \mathcal{C}(Y^{(i)}, \;\hat{Y}^{(\sigma (i))}) \label{eqn:matching}
\end{equation}

where $\hat{Y}^{(j)} = (\hat{\mathbf{P}}^{(j)}, \;\hat{\mathbf{C}}^{(j)})$ is the prediction to be matched and $Y^{(i)}$ is the ground truth. For simplicity, we use the control point location to guide the learning of character decoding. Therefore, the matching cost is defined as a mixture of confidence and coordinate deviation. For $i$-th ground truth and its matched $\sigma(i)$-th query, the cost function is 
\begin{equation}
    \resizebox{.9\linewidth}{!}{%
    $\mathcal{C}(Y^{(i)}, \hat{Y}^{(\sigma (i))}) = \lambda_{\text{cls}} \text{FL}'(\hat{b}^{(\sigma(i))}) + \lambda_{\text{coord}} \sum_{k=1}^N \left\| p_k^{(i)} - \hat{p}_k^{(\sigma(i))}\right\|$
} \label{eqn:matching_cost}
\end{equation}

where $\hat{b}^{(\sigma(i))}$ is the probability for the only instance class -- text, which also serves as the confidence score. $\text{FL}'$ is derived from the focal loss\cite{Lin_2017_ICCV}, and is defined as the difference of the positive and negative term: $\text{FL}'(x)=-\alpha (1-x)^\gamma \log(x) + (1-\alpha) x^\gamma \log(1-x)$. The second term in Equation \ref{eqn:matching_cost} is the L-1 distance between ground truth and predicted control point coordinates. 

The problem in \ref{eqn:matching} can be efficiently solved by the Hungarian algorithm\cite{1955_hungarian}. We use the same bipartite matching scheme to match proposals in the guidance generator with ground truth boxes, which are bounding boxes for the control points.

\vspace{.5em}

\textbf{Instance classification loss.} We adopt focal loss as the classification loss of text instances. For the $j$-th query, the loss is defined as:
\begin{equation}
\begin{aligned}
    \mathcal{L}_{\text{cls}}^{(j)} = & -\mathds{1}_{\left\{ j \in \text{Im}(\sigma) \right\}} \alpha (1-\hat{b}^{(j)})^\gamma \log (\hat{b}^{(j)}) \\
    & -\mathds{1}_{\left\{ j \notin \text{Im}(\sigma) \right\}} (1-\alpha) (\hat{b}^{(j)})^\gamma \log (1-\hat{b}^{(j)}) 
\end{aligned}
\end{equation}

where $\text{Im}(\sigma)$ is the image of the mapping $\sigma$.

\vspace{.5em}

\textbf{Control point loss.} L-1 distance loss is used for control point coordinate regression:
\begin{equation}
    \mathcal{L}_{\text{coord}}^{(j)} = \mathds{1}_{\left\{ j \in \text{Im}(\sigma) \right\}} \sum_{i=1}^N \left\|p_i^{(\sigma^{-1}(j))} - \hat{p}_i^{(j)} \right\|
\end{equation}

\vspace{.5em}

\textbf{Character classification loss.} We deem the character recognition as a classification problem, where each class is assigned a specific character. Cross entropy loss is used here:
\begin{equation}
    \mathcal{L}_{\text{char}}^{(j)} = \mathds{1}_{\left\{ j \in \text{Im}(\sigma) \right\}} \sum_{i=1}^M \left( -c_i^{(\sigma^{-1}(j))} \log  \hat{c}_i^{(j)} \right)
\end{equation}

The loss function for the dual decoders comprises the three aforementioned losses:
\begin{equation}
    \mathcal{L}_{\text{dec}} = \sum_{j} \left( \lambda_{\text{cls}} \mathcal{L}_{\text{cls}}^{(j)}  + \lambda_{\text{coord}} \mathcal{L}_{\text{coord}}^{(j)} + \lambda_{\text{char}} \mathcal{L}_{\text{char}}^{(j)} \right)
\end{equation}

\vspace{.5em}

\textbf{Bounding box intermediate supervision loss.} To make the proposals in Section \ref{sec:box_to_poly} more accurate, we also introduce intermediate supervision for them at the encoder side. The same bipartite matching scheme is used to match these bounding box proposals to the ground truth. We denote the matching here as $\sigma'$, and the overall loss here is 
\begin{equation}
    \mathcal{L}_\text{enc} = \sum_{i} \left( \lambda_{\text{cls}}\mathcal{L}_{\text{cls}}^{(i)} + \lambda_{\text{coord}}\mathcal{L}_{\text{coord}}^{(i)} + \lambda_{\text{gIoU}}\mathcal{L}_{\text{gIoU}}^{(i)} \right)
\end{equation}

where the classification loss $\mathcal{L}_\text{cls}^{(i)}$ and control point loss $\mathcal{L}_{\text{coord}}^{(i)}$ are identical to the ones used for polygon detection, except for the different matching $\sigma'$ used. $\mathcal{L}_{\text{gIoU}}$ is the generalized IoU loss defined in \cite{rezatofighi2019generalized} for bounding box regression.

The final loss for the entire model is simply the sum of the encoder and decoder loss.

\section{Experiments}

\subsection{Datasets}

Here we briefly introduce the datasets used in this paper.

\vspace{.5em}

\textbf{SynthText 150k.} Unlike existing \textit{SynthText 800k} which contains mostly straight texts in quadrilateral annotations, SynthText 150k synthesized in \cite{liu2020abcnet} comes with 94,723 images containing mostly straight text and 54,327 with major curved texts in Bezier annotations. 

\vspace{.5em}

\textbf{ICDAR 2015.} The ICDAR 2015\cite{karatzas2015icdar} is the official dataset for ICDAR 2015 Robust Reading Competition. It contains 1000 training images and 500 testing images, with horizontal and perspective texts with quadrilateral box annotation. The images were captured with hand-held cameras in the wild, therefore blurs and obscurities are frequent.  

\vspace{.5em}

\textbf{Total-Text.} The Total-Text\cite{Total-Text} is a popular curved text benchmark, with 1255 images for training and 300 for testing. Word-level polygon or Bezier annotations are used. 

\vspace{.5em}

\textbf{CTW1500.} \cite{ctw-1500} is another important curved scene text benchmark, with 1000 training images and 500 testing images. Different from Total-Text, it contains both English and Chinese texts. As the proportion of Chinese texts is small, we ignore them during training.

We follow the standard evaluation protocols used in these datasets, which involve the calculation of IoU between predicted and ground truth polygons. The output of TESTR with Bezier annotations is converted back to polygons prior to evaluation. 

\begin{table*}[!ht]
\centering
\caption{\small Scene text spotting results on Total-Text. ``None'' refers to recognition without lexicon. ``Full'' lexicon contains all the words in the test set.}\label{tab:total_text}
\vspace{-0.3em}
\setlength{\tabcolsep}{10pt}
\resizebox{.8\linewidth}{!}{%
\begin{tabular}{ll ccc cc c}
\toprule
\multirow{2}{*}{Method} & \multirow{2}{*}{Backbone} & \multicolumn{3}{c}{Detection} & \multicolumn{2}{c}{End-to-End} & \multirow{2}{*}{FPS} \\
\cmidrule(lr){3-5} \cmidrule(lr){6-7}
& & P & R & F & None & Full &  \\
\midrule
FOTS\cite{liu2018fots} & ResNet-50 & 52.3 & 38.0 & 44.0 & 32.2 & $-$ & $-$ \\ 
Textboxes\cite{liao2016textboxes} & ResNet-50-FPN & 62.1 & 45.5 & 52.5 & 36.3 & 48.9 & 1.4 \\
Mask TextSpotter\cite{lyu2018mask} & ResNet-50-FPN & 69.0 & 55.0 &  61.3 &  52.9 & 71.8 & 4.8 \\
CharNet\cite{xing2019convolutional} & ResNet-50-Hourglass57 & 87.3 & 85.0  & 86.1 & 66.2 & $-$ & 1.2 \\
Text Dragon\cite{feng2019textdragon}  & VGG16 & 85.6 & 75.7 & 80.3 & 48.8 & 74.8 & $-$ \\
Boundary TextSpotter\cite{Wang_Lu_2020_boundary} & ResNet-50-FPN & 88.9 & 85.0 & 87.0 & 65.0 & 76.1 & $-$ \\
Unconstrained\cite{qin2019unconstrained} & ResNet-50-MSF & 83.3 & 83.4 & 83.3 & 67.8 & $-$ & $-$  \\
Text Perceptron\cite{qiao2021text} & ResNet-50-FPN  & 88.8 & 81.8 & 85.2 & 69.7 & 78.3 & $-$ \\
Mask TextSpotter v3\cite{liao2020mask}  & ResNet-50-FPN & $-$ & $-$ & $-$ & 71.2 & 78.4 & $-$ \\ 
ABCNet-MS\cite{liu2020abcnet} & ResNet-50-FPN & $-$ & $-$ & $-$  & 69.5 & 78.4 & 6.9 \\
ABCNet v2\cite{liu2021abcnet} & ResNet-50-FPN & 90.2 & 84.1  & 87.0 & 70.4 & 78.1 & 10 \\ 
MANGO\cite{qiao2021mango} &  ResNet-50-FPN & $-$ & $-$ & $-$ & 72.9 & 83.6 & 4.3 \\
PGNet\cite{Wang_2021_pgnet} & ResNet-50-FPN & 85.5 & \textbf{86.8}   & 86.1 & 63.1 & $-$ & 35.5 \\
\midrule
TESTR-Bezier (ours) & ResNet-50 & 92.8 & 83.7 & \textbf{88.0} & 71.6 & 83.3 & 5.5 \\
TESTR-Polygon (ours) & ResNet-50 & \textbf{93.4} & 81.4 & 86.9 & \textbf{73.3} & \textbf{83.9} & 5.3 \\
\bottomrule
\end{tabular}
}
\end{table*}

\subsection{Implementation Details}

We use ResNet-50\cite{he2016deep} as the feature backbone for all the experiments. Multi-scale feature maps are directly drawn from the last three stages of ResNet without FPN. The parameters for the deformable Transformers are similar to  \cite{zhu2021deformable}, with $H=8$ heads and $K=4$ sampling points for the deformable attentions, and we use 6 layers of encoders and decoders. 

\vspace{.5em}

\textbf{Data augmentation.} The data augmentation during training is conducted by 1) random resize with the shorter edge ranging from 480 to 896, and the longest edge kept within 1600; 2) instance-aware random crop, which ensures the cropped size larger than half of the original size and no texts being cut. During test time, we resize the shorter edge to 1600 while keeping the longest edge within 1892. 

\vspace{.5em}

\textbf{Pre-training.} The model is pretrained on a mixture of SynthText 150k, MLT 2017\cite{nayef2019icdar2019} and TotalText for 440k iterations. The base learning rate for the polygon variant is $1 \times 10^{-4}$ and is decayed at the 340k-th iteration by a factor of 0.1. Learning rates are
scaled by a factor of 0.1 for the linear projections used to predict reference points, sampling offsets of the multi-scale deformable attention and feature backbone. AdamW\cite{Loshchilov2019DecoupledWD} is used as the optimizer, with $\beta_1=0.9$, $\beta_2=0.999$ and weight decay of $10^{-4}$. We use $Q=100$ composite queries. The max text length $M$ is $25$, and number of polygon control points $N$ is 16. The weighting factors for the losses are $\lambda_{\text{cls}}=2.0$, $\lambda_{\text{coord}}=5.0$, $\lambda_{\text{char}}=4.0$, $\lambda_{\text{gIoU}}=2.0$. We set $\alpha=0.25$, $\gamma=2.0$ for the focal loss. For the Bezier variant of the model, we have $N=8$ control points, double the value of base learning rate, and half $\lambda_{\text{char}}$ for the purpose of balancing. The pre-training process takes about 3 days on 8 RTX 2080Ti GPUs with the image batch size of 8.

\vspace{.5em}

\textbf{Finetuning.} The model is finetuned on specific datasets prior to evaluation to mitigate the variance across different datasets. For the Total-Text and ICDAR 2015 dataset, we finetune the model for 20k iterations, with the base learning rate scaled by 0.1. For CTW1500, to address the longer texts present in the dataset, the maximum text length $M$ is set to 100, and therefore the model is finetuned for 200k iterations, larger than the ones needed for the other two.

\subsection{Results}
\begin{figure*}[!ht]
    \centering
    \includegraphics[width=\linewidth]{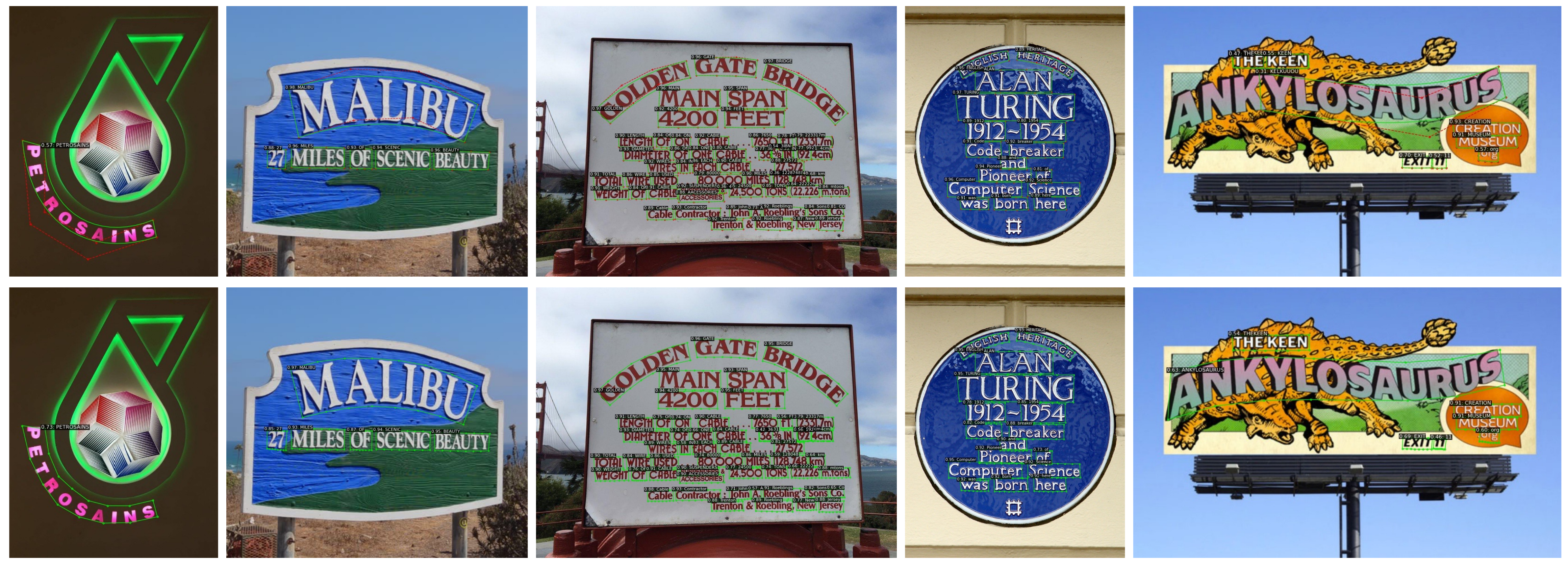}
    \caption{\small Qualitative results on Total-Text without lexicons. Top row: Bezier; bottom row: polygon annotations. The predictions are shown in green contours, with Bezier control points in red. The number before text is the confidence score. TESTR-Bezier fails to capture the shape of the ``ANKYLOSAURUS'' text in the last column, while the polygon variant succeeds. Zoom in for better visualization. }
    \label{fig:qualitative_totaltext}
    \vspace{-.5em}
\end{figure*}

Here we present the benchmark of our model TESTR in polygonal or Bezier curve annotations. 

\begin{figure}[!ht]
    \centering
    \includegraphics[width=\linewidth,height=1.5\linewidth]{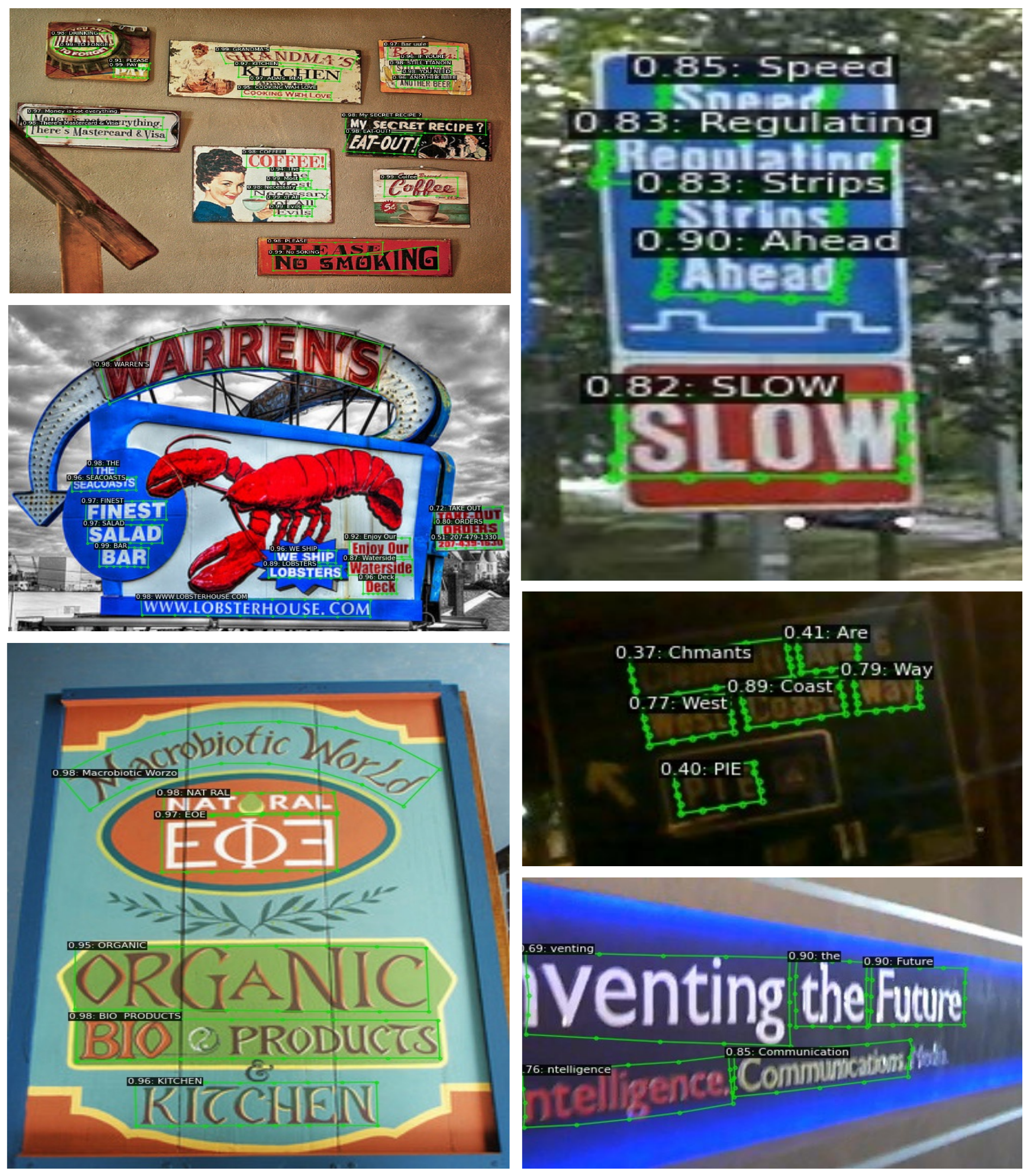}
    \vspace{-2em}
    \caption{\small Qualitative results of TESTR on CTW1500 (left column) and ICDAR (right column) using polygonal annotations.}
    \label{fig:qualitative_ctw_icdar}
    \vspace{-1em}
\end{figure}

\begin{table}[!htb]
    \centering
    \caption{\small End-to-end text spotting results on CTW1500. ``None'' represents lexicon-free, while ``Full'' indicates all the words in the test set are used.}\label{tab:ctw1500}
    \vspace{-0.3em}
\resizebox{.9\linewidth}{!}{%
    \begin{tabular}{l ccc cc}
    \toprule
    \multirow{2}{*}{Method} & \multicolumn{3}{c}{Detection} & \multicolumn{2}{c}{End-to-End} \\ \cmidrule(lr){2-4} \cmidrule(lr){5-6} 
     & P & R & F & None & Full  \\ 
     \midrule
    Text Dragon\cite{feng2019textdragon} & 84.5 & 82.8 & 83.6 & 39.7 & 72.4 \\ 
    Text Perceptron\cite{qiao2021text} & 87.5 & 81.9 & 84.6 & 57.0 & $-$  \\
    ABCNet\cite{liu2020abcnet} & $-$ & $-$ & $-$ & 45.2 & 74.1  \\ 
    ABCNet v2\cite{liu2021abcnet} & 85.6 & \textbf{83.8} & 84.7 &  57.5 & 77.2 \\
    MANGO\cite{qiao2021mango} & $-$ & $-$ & $-$ & \textbf{58.9} & 78.7  \\
    \hline
    TESTR-Bezier (ours) & 89.7 & 83.1 & 86.3 & 53.3 & 79.9  \\ 
    TESTR-Polygon (ours) & \textbf{92.0} & 82.6 & \textbf{87.1} & 56.0 & \textbf{81.5} \\
    \bottomrule
    \end{tabular}%
}
\vspace{-1.5em}
\end{table}

\vspace{.5em}

\textbf{Irregular texts.} We test our method on two irregular text benchmarks: Total-Text and CTW1500, and the quantitative results are shown in Table \ref{tab:total_text} and \ref{tab:ctw1500}.

In terms of text detection, the TESTR-Beizer outperforms the previous most accurate model by 1.0\% on the F-score metric on the Total-Text dataset. The TESTR-Polygon has almost the same detection accuracy as ABCNet v2 and is free of Bezier annotations. On the CTW-1500 dataset, the F-score of TESTR surpasses that of ABCNet v2 by a large margin, with 1.6\% for Bezier and 2.4\% for polygonal annotations. 

In the case of end-to-end text spotting, TESTR-Polygon significantly surpasses the best-reported results by 2.8\% when equipped with full lexicons on CTW1500. On Total-Text, our method outperforms the previous best results by 0.4\% without lexicons and by 0.3\% with full lexicons.

Qualitative results on the two datasets are shown in the Figure \ref{fig:qualitative_totaltext} and \ref{fig:qualitative_ctw_icdar}. The results illustrate our method can handle both straight and curved texts well. The failure cases for TESTR with Bezier annotations are displayed, \eg the last column of Figure \ref{tab:total_text}, where it fails to generate the correct bounding polygon for the Bezier curves, while the polygon model variant succeeds. This observation is consistent with the quantitative results.

In summary, the results on Total-Text and CTW1500 demonstrate the effectiveness of our method for arbitrarily-shaped text spotting. Meanwhile, the overall performance of TESTR-Polygon is better than TESTR-Bezier mostly.

\textbf{Regular texts.} We evaluate our method on ICDAR2015 containing many perspective texts annotated with quadrilateral bounding boxes, and the results are shown in Table \ref{tab:icdar2015}. In the detection stage, our method achieves state-of-the-art F-score. In the end-to-end text spotting, our method exhibits remarkable performance in the lexicon-free setting, on par with Text Perceptron with generic lexicons. When lexicons are available, TESTR works best with the ``Strong'' type, obtaining competitive results compared with other methods. Qualitative results in the right column of Figure \ref{fig:qualitative_ctw_icdar} show our method can recognize texts even in occluded scenes or from extreme viewing angles. 

\begin{table}[!htb]
\caption{\small Results on ICDAR 2015 dataset. ``S'', ``W'', ``G'', ``N'' represent recognition with ``Strong'', ``Weak'', ``Generic'' or ``None'' lexicon respectively.}\label{tab:icdar2015}
\centering
\setlength{\tabcolsep}{3pt}
\resizebox{\linewidth}{!}{%
\begin{tabular}{l ccc ccc c}
\toprule
\multirow{2}{*}{Method} & \multicolumn{3}{c}{Detection} & \multicolumn{4}{c}{End-to-End} \\
\cmidrule(lr){2-4} \cmidrule(lr){5-8}
 & P & R & F & S & W & G & N \\
\midrule
He \textit{et al.}\cite{He2018AnET} & 87.0 & 86.0 & 87.0 & 82.0 & 77.0 & 63.0 & $-$ \\
TextNet\cite{sun2018textnet} & 89.4 & 85.4 & 87.4 & 78.7 & 74.9 & 60.5 & $-$ \\ 
FOTS\cite{liu2018fots} & 91.0 & 85.2 & 88.0 & 81.1 & 75.9 & 60.8 & $-$ \\
CharNet R-50\cite{xing2019convolutional} & 91.2 & 88.3 & 89.7 & 80.1 & 74.5 & 62.2 & 60.7 \\
Boundary TextSpotter\cite{Wang_Lu_2020_boundary} & 89.8 & 87.5 & 88.6 & 79.7 & 75.2 & 64.1 & $-$ \\
Unconstrained\cite{qin2019unconstrained} & 89.4 & 85.8 & 87.5 & 83.4 & \textbf{79.9} & 68.0 & $-$ \\
Text Perceptron\cite{qiao2021text} & \textbf{92.3} & 82.5 & 87.1 & 80.5 & 76.6 & 65.1 & $-$ \\
Mask TextSpotter v3\cite{liao2020mask} & $-$ & $-$ & $-$ & 83.3 & 78.1 & \textbf{74.2} & $-$ \\
ABCNet v2\cite{liu2021abcnet} &  90.4 & 86.0 & 88.1 & 82.7 & 78.5 & 73.0 & $-$  \\
MANGO\cite{qiao2021mango} & $-$ & $-$ & $-$ & 81.8 & 78.9 & 67.3 & $-$ \\
PGNet\cite{Wang_2021_pgnet} & 91.8 & 84.8 & 88.2 & 83.3 & 78.3 & 63.5 & $-$ \\
\midrule
TESTR-Polygon (ours) & 90.3 & \textbf{89.7} & \textbf{90.0} & \textbf{85.2} & 79.4 & 73.6 & \textbf{65.3}  \\
\bottomrule
\end{tabular}
}
\end{table}

\subsection{Ablation Studies} \label{sec:ablations}

To illustrate the effectiveness of the proposed components, we conduct multiple ablation studies on Total-Text with polygonal annotations.

\vspace{.5em}

\textbf{Box-to-polygon detection process.} In our design of TESTR, the encoder performs multi-scale self-attention across feature maps, and a guidance generator produces coarse bounding boxes from the encoded features. These bounding boxes, encoded and added on top of the learnable control point query embeddings, are used to guide the learning of control point regression in the location decoder. We ablate this module by replacing the $\varphi(w^{(i)})$ term in Equation \ref{eqn:box_guidance} with a learnable embedding vector to show how the bounding box guidance affects the results. The results shown in Table \ref{tab:ablation_modules} demonstrate that the box-to-polygon detection process could improve Precision, Recall, F-score by 0.5\%, 3.6\% and 2.2\% in detection respectively, and significantly improve the end-to-end recognition results by 5.8\%.

\vspace{.5em}

\textbf{Multi-scale feature.} Our method leverages multi-scale feature maps to overcome the challenge of the prevalent small text instances in the images. We conduct ablations by using only the feature map from the last stage of ResNet. Table \ref{tab:ablation_modules} shows that adopting multi-scale features could improve Precision, Recall, F-score by 1.2\%, 2.3\% and 1.8\% in detection respectively, and dramatically improve the end-to-end results by 10.8\%. This indicates the text recognition task benefits much from features with larger scales.

\begin{table}[!htb]
\centering
\caption{\small Ablation study on Total-Text using TESTR with polygonal output. }\label{tab:ablation_modules}
\setlength{\tabcolsep}{4pt}
\resizebox{.9\linewidth}{!}{%
\begin{tabular}{cc ccc c}
\toprule
 \multirow{2}{*}{Multi-scale Features} & \multirow{2}{*}{Box Guidance}& \multicolumn{3}{c}{Detection} & \multirow{2}{*}{E2E} \\
  \cmidrule(lr){3-5}
 &    & P & R & F &  \\
 \midrule
 $-$ & \checkmark & 92.2 & 79.1 & 85.1 &  62.5  \\
\checkmark & $-$ & 92.9 & 77.8 & 84.7 &  67.5  \\
\checkmark & \checkmark & \textbf{93.4} & \textbf{81.4} & \textbf{86.9} & \textbf{73.3} \\
 \bottomrule
\end{tabular}%
}
\end{table}

\textbf{Input scale.} To demonstrate the tradeoffs between speed and accuracy, we evaluate our model with the shorter side of the image resized to 720, 1000, 1280, 1600 respectively. The results are shown in Table \ref{tab:ablation_input_scale}. The F-score of both detection and end-to-end recognition increases with FPS decreasing as the input scale grows larger.

\begin{table}[!htb]
\centering
\caption{\small Performance of TESTR with different input scales on Total-Text.}\label{tab:ablation_input_scale}
\setlength{\tabcolsep}{8pt}
\resizebox{.9\linewidth}{!}{%
\begin{tabular}{cc ccc c c}
\toprule
\multirow{2}{*}{Model Type}& \multirow{2}{*}{Input} & \multicolumn{3}{c}{Detection} & \multirow{2}{*}{E2E} &  \multirow{2}{*}{FPS} \\
  \cmidrule(lr){3-5}
    & & P & R & F & & \\
 \midrule
 \multirow{4}{*}{Bezier} & 720 & 91.5 & 81.5 & 86.2 & 62.6 & 11.6 \\
  & 1000 & 91.2 & \textbf{84.1} & 87.5 & 69.4 & 7.9 \\
  & 1280 & 92.3 & 83.7 & 87.8 & 70.9 & 5.8 \\
  & 1600 & \textbf{92.8} & 83.7 & \textbf{88.0} & \textbf{71.6} & 5.5 \\
 \midrule
 \multirow{4}{*}{Polygon} & 720 & 92.7 & 79.7 & 85.7 & 66.2 & 11.7 \\
  & 1000 & 92.1 & 81.4 & 86.4 & 70.5 & 8.0\\
  & 1280 & 92.5 & \textbf{81.5} & 86.7 & 72.2 & 6.0 \\
  & 1600 & \textbf{93.4} & 81.4 & \textbf{86.9} & \textbf{73.3} & 5.3 \\
 \bottomrule
\end{tabular}%
}
\end{table}

\vspace{.5em}

\textbf{Balancing between dual decoders.} The decoder loss $\mathcal{L}_{\text{dec}}$ comprises loss functions for the location and character decoder respectively. We ablate on $\lambda_{\text{char}}$ to demonstrate the effects of the balancing between the two decoders. The results in Table \ref{tab:ablation_w_char} show our method works well with $\lambda_{\text{char}}$ in a wide range $4.0$ - $10.0$, while it performs best with $\lambda_\text{char}=8.0$. We choose $\lambda_\text{char}=4.0$ in the main experiments to avoid extensive hyperparameter tuning.

\begin{table}
\centering
\setlength{\tabcolsep}{8pt}
\caption{\small TESTR-Polygon on TotalText with different $\lambda_{\text{char}}$.}
\label{tab:ablation_w_char}
\resizebox{.8\linewidth}{!}{%
\begin{tabular}{c ccc ccc }
\toprule
\multirow{2}{*}{$\lambda_{\text{char}}$} & \multicolumn{3}{c}{Detection} & \multicolumn{3}{c}{End-to-End} \\
\cmidrule(lr){2-4} \cmidrule(lr){5-7}
 & P & R & F & P & R & F \\
\midrule
1.0 & 93.3 & 80.7 & 86.5 & 75.4 & 68.1 & 71.6 \\
2.0 & 93.3 & 81.8 & 87.2 & 75.7 & 69.2 & 72.3 \\
4.0 & \textbf{93.4} & 81.4 & 86.9 & \textbf{76.9} & 70.0 & 73.3 \\
6.0 & 93.3 & 81.2 & 86.8 & 76.8 & 69.8 & 73.1 \\
8.0 & 92.2 & \textbf{82.9} & \textbf{87.3} & 76.7 & \textbf{71.1} & \textbf{73.8} \\
10.0 & 92.7 & 81.4 & 86.7 & \textbf{76.9} & 70.1 & 73.4 \\
\bottomrule
\end{tabular}
}
\end{table}

\section{Discussions}

\textbf{Limitations and future work} ~ In our setting of TESTR, we assume a fixed number of polygon control points, which might not be optimal. For most perspective texts, quadrilaterals would be sufficient, while many more vertices would be required if texts come with higher curvature. In the future, we would like to investigate methods that adaptively determine the adequate number of polygon control points within our framework to better capture their shapes.

\vspace{.5em}

\textbf{Conclusions}~ In this paper, we have presented TESTR, a text spotting framework based on single-encoder dual-decoder Transformer architecture. By modeling the text detection and recognition in a holistic fashion, our model directly performs set prediction without heuristics-driven post-processing or Region-of-Interest operations. A bounding-box guided polygon detection procedure allows efficient detection of arbitrarily-shaped texts. In addition, our canonical representation of control points enables the model to function effectively for both polygonal and Bezier annotations. Experimental results on challenging curved or oriented text benchmarks, Total-Text and CTW1500, demonstrate the state-of-the-art performance of TESTR.

\vspace{.5em}

\textbf{Acknowledgement} We thank Intel Corporation for an award. We thank Weijian Xu, Yifan Xu, and Tianyi Xiong for valuable feedbacks.

{\small
\bibliographystyle{ieee_fullname}
\bibliography{main}
}

\end{document}